\documentclass[sigconf,natbib=true]{acmart}

\usepackage{makecell}
\usepackage{amsmath}
\usepackage{graphicx}
\usepackage{epstopdf} 
\usepackage{multirow}
\usepackage{bm}
\usepackage{times}
\usepackage{latexsym}
\usepackage{amsfonts}
\usepackage{microtype}
\usepackage{color}
\AtBeginDocument{%
  }

\usepackage{graphicx}               
\usepackage{tabularx}               
\usepackage{booktabs}
\usepackage{amsmath}
\usepackage{stmaryrd}
\usepackage{xcolor}
\usepackage{soul}

\newcolumntype{C}{>{\centering\arraybackslash}X}
\usepackage{multirow}               
\usepackage{diagbox}                
\usepackage{hhline}                 
\usepackage{color}                  
\usepackage{amsmath}                
\usepackage{mathtools}              
\usepackage{enumitem}               

\usepackage{subfigure}
\usepackage{booktabs}
\usepackage{extarrows}
\usepackage{makecell}

\usepackage{soul}

\definecolor{RoseQuartzBg}{HTML}{F7CAC9}
\definecolor{RoseQuartz}{HTML}{F5A798}
\definecolor{Serenity}{HTML}{92A8D1}
\definecolor{OrangeRed}{rgb}{1.0, 0.27, 0.0}
\definecolor{Red}{rgb}{1.0, 0.0, 0.0}
\definecolor{Turquoise}{HTML}{0F4C81}
\usepackage{xparse}
\NewDocumentCommand{\lifu}{ mO{} }{\textcolor{OrangeRed}{\textsuperscript{\textit{Lifu}}\textsf{\textbf{\small[#1]}}}}
\NewDocumentCommand{\minqian}{ mO{} }{\textcolor{blue}{\textsuperscript{\textit{Minqian}}\textsf{\textbf{\small[#1]}}}}
\NewDocumentCommand{\zhiyang}{ mO{} }{\textcolor{Serenity}{\textsuperscript{\textit{Zhiyang}}\textsf{\textbf{\small[#1]}}}}

\NewDocumentCommand{\jy}{ mO{} }{\textcolor{Red}{\textsuperscript{\textit{jy}}\textsf{\textbf{\small[#1]}}}}

\setcopyright{acmcopyright}

\copyrightyear{2023}
\acmYear{2023}
\setcopyright{rightsretained}
\acmConference[SIGIR '23]{Proceedings of the 46th International ACM SIGIR
Conference on Research and Development in Information Retrieval}{July 23--27,
2023}{Taipei, Taiwan}
\acmBooktitle{Proceedings of the 46th International ACM SIGIR Conference on
Research and Development in Information Retrieval (SIGIR '23), July 23--27,
2023, Taipei, Taiwan}\acmDOI{10.1145/3539618.3591781}
\acmISBN{978-1-4503-9408-6/23/07}

\begin{document}

\title{Understand the Dynamic World: An End-to-End Knowledge Informed Framework for Open Domain Entity State Tracking}





\author{Mingchen Li}
\affiliation{%
  \institution{Georgia Sate University}
  \city{Atlanta}
  \country{USA}}
\email{lmingchen96@gmail.com}

\author{Lifu Huang}
\affiliation{%
  \institution{Virginia Tech}
  \city{Blacksburg}
  \country{USA}
}
\email{lifuh@vt.edu}

\begin{abstract}

Open domain entity state tracking aims to predict reasonable state changes of entities (i.e., [\textit{attribute}] of [\textit{entity}] was [\textit{before\_state}] and [\textit{after\_state}] afterwards) given the action descriptions. It's important to many reasoning tasks to support human everyday activities. 
However, it's challenging as the model needs to predict an arbitrary number of entity state changes caused by the action while most of the entities are implicitly relevant to the actions and their attributes as well as states are from open vocabularies. 
To tackle these challenges, we propose a novel end-to-end \textbf{K}nowledge \textbf{I}nformed framework for open domain \textbf{E}ntity \textbf{S}tate \textbf{T}racking, namely \textsc{Kiest}, which explicitly retrieves the relevant entities and attributes from external knowledge graph (i.e., ConceptNet) and incorporates them to autoregressively generate all the entity state changes with a novel dynamic knowledge grained encoder-decoder framework. To enforce the logical coherence among the predicted entities, attributes, and states, we design a new constraint decoding strategy and employ a coherence reward to improve the decoding process. Experimental results show that our proposed \textsc{Kiest} framework significantly outperforms the strong baselines on the public benchmark dataset -- OpenPI.\footnote{The source code, data, and checkpoints will be publicly available at \url{https://github.com/VT-NLP/open_domain_entity_state_tracking}.}

\end{abstract}

\ccsdesc[500]{Computing methodologies~Natural language processing}
\keywords{Open Domain Entity State Tracking, Knowledge Informed Generation, Constraint Decoding, Coherent Reward}

\maketitle
\section{Introduction} \label{intro}

\begin{figure}[t]
	\centering
	\includegraphics[width=0.95\columnwidth]{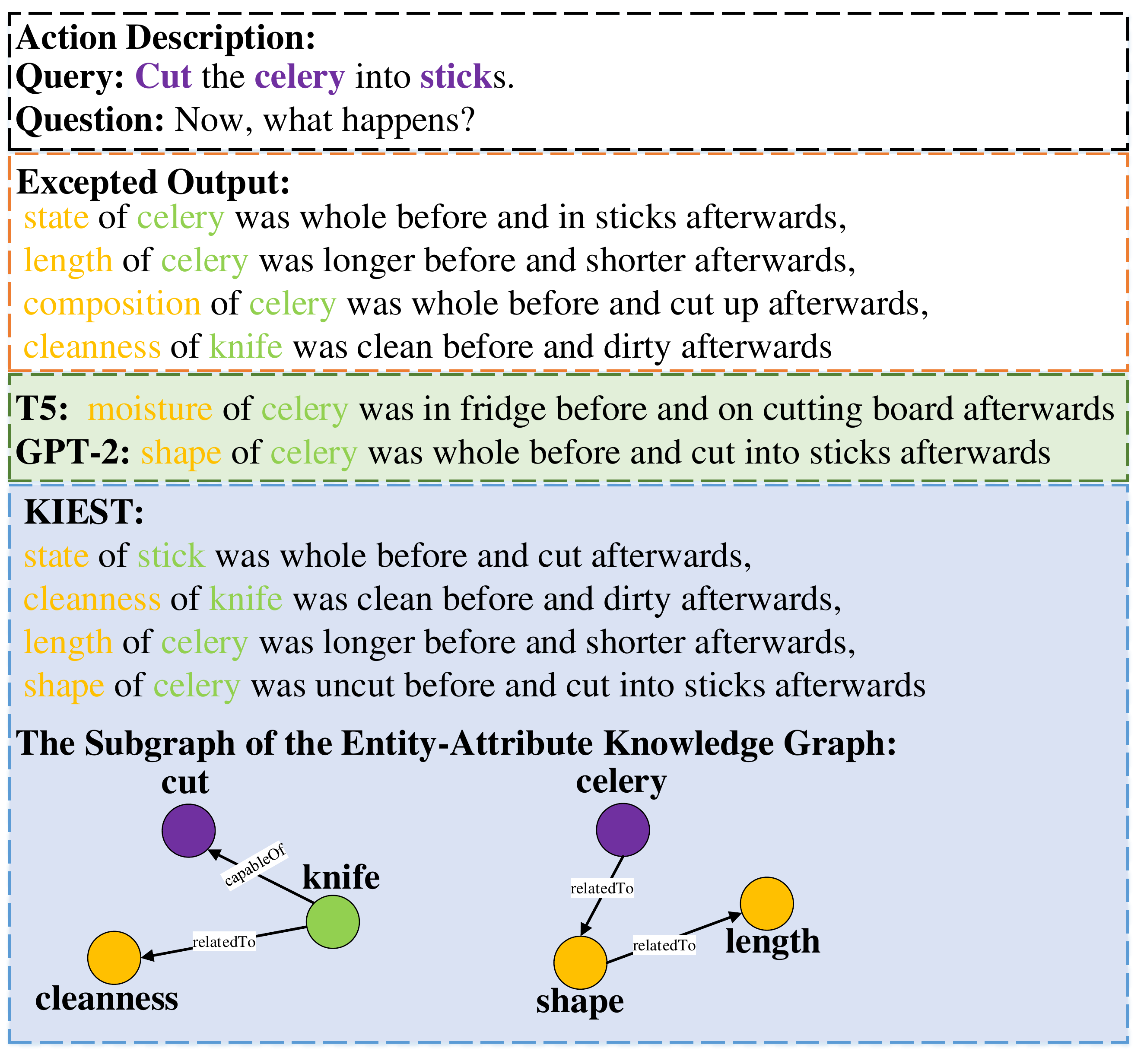}
	\caption{An example demonstrating the input,  expected outputs and system outputs (from T5~\cite{raffel2020exploring}, GPT-2~\cite{radford2019language} and our \textsc{Kiest} approach)   for open-domain entity state tracking. The purple words in the action description highlight the keywords that are used to retrieve the additional entity and attribute knowledge, e.g., the concepts shown in the bottom entity-attribute knowledge graph. The green and orange words in the expected outputs and system outputs highlight the entities and attributes for describing the entity state changes. 
 }
	\label{con:subgraph}
\end{figure}

Open domain entity state tracking~\cite{tandon2020dataset} aims to predict all the state changes of entities that are related to a given action description, where each state change can be described with a template, e.g., ``[\textit{attribute}] of [\textit{entity}] was [\textit{before\_state}] and [\textit{after\_state}] afterwards''.
It's an important task to many reasoning and information retrieval systems to better understand human daily activities and make recommendations, e.g., the subsequent actions that humans need to perform. 
Figure~\ref{con:subgraph} shows an example, where given the action, \textit{Cut the celery into sticks.}, we need to infer all the entity state changes that are related to the action, such as \textit{length of celery was longer before and shorter afterwards}, \textit{cleanness of knife was clean before and dirty afterwards}, and so on.

There are two particular challenges for the open domain entity state tracking task: first, given that the actions are from open domains, the entities, attributes, and states that are related to state changes are usually from open set vocabularies, e.g., in the example of Figure~\ref{con:subgraph},  the entity such as \textit{knife} is not explicitly mentioned in the action description and there is no close-set attribute vocabulary to indicate the state changes of entities,  
making it hard to be formulated as a reading comprehension task as previous studies~\cite{mishra2018tracking,faghihi2021time,spiliopoulou2022events} which only focus on a few predefined entities and states (e.g., \textit{location}, \textit{existence}).
In addition, there could be an arbitrary number of entity state changes that are caused by the action while the system is expected to predict most, if not all, state changes of entities. 
Previous study~\cite{tandon2020dataset} tackles these challenges by exploring an autoregressive language model, such as GPT-2~\cite{radford2019language}, to directly generate all the entity state changes (green box in Figure~\ref{con:subgraph}). 
However, as shown in Figure~\ref{con:subgraph}, these approaches suffer from very low coverage of the entities and attributes for the predicted state changes, and without any constraint, the models easily generate state changes that are not related to the context of the action or consistent with human commonsense, e.g., the generated ``\textit{fridge}'' (\textit{before\_state}) from T5 is not coherent with the attribute ``\textit{moisture}''.



In this work, we argue that knowledge graphs (KGs), such as ConceptNet~\cite{liu2004conceptnet}, can provide meaningful knowledge to inform the model of relevant entities and attributes to the given action description, and thus facilitate the model to better predict entity state changes. For example, as shown in Figure~\ref{con:subgraph}, given the source concepts, such as \textit{cut}, from the action description, ConceptNet provides more implicitly relevant concepts, such as \textit{knife}, \textit{cleanness} and so on.  
Motivated by this, we propose a novel end-to-end \textbf{K}nowledge \textbf{I}nformed framework for open domain \textbf{E}ntity \textbf{S}tate \textbf{T}racking, namely \textsc{Kiest}, which consists of two major steps: (1) retrieving and selecting all the relevant entities and attributes to the given action description from ConceptNet; (2) incorporating the external entity and attribute knowledge into a novel \textit{Dynamic Knowledge Grained Encoder-Decoder} framework, which is based on an autoregressive pre-trained language model, such as T5~\cite{raffel2020exploring}, to generate all the entity state changes. To encourage the model to generate coherent and reasonable state changes, we further propose a constrained decoding strategy and an \textsl{\textbf{E}ntity \textbf{S}tate \textbf{C}hange  \textbf{C}oherence  \textbf{R}eward (ESCCR)} to improve the generation process. The experimental results demonstrate the effectiveness of our proposed \textsc{Kiest} framework with a significant improvement over the strong baselines. The contributions of this work can be summarized as:

\begin{itemize}
     \item To the best of our knowledge, we are the first to incorporate the external entity and attribute knowledge to inform the model to better generate entity state changes with higher coverage;

     \item We design a novel Dynamic Knowledge Grained Encoder-Decoder approach to dynamically incorporate the external knowledge and autoregressively generate all the entity state changes; 
     \item We also design a new constrained decoding strategy and an automatic reward function to estimate the coherence of entity state changes so as to encourage the model to generate state changes that are more coherent to the context actions and human commonsense;
    \item  We conduct a thorough analysis of our method, including an ablation study, demonstrating the robustness of our framework.
\end{itemize}

\section{Related Work}

\begin{figure*}[t]
	\centering
	\includegraphics[width=2\columnwidth]{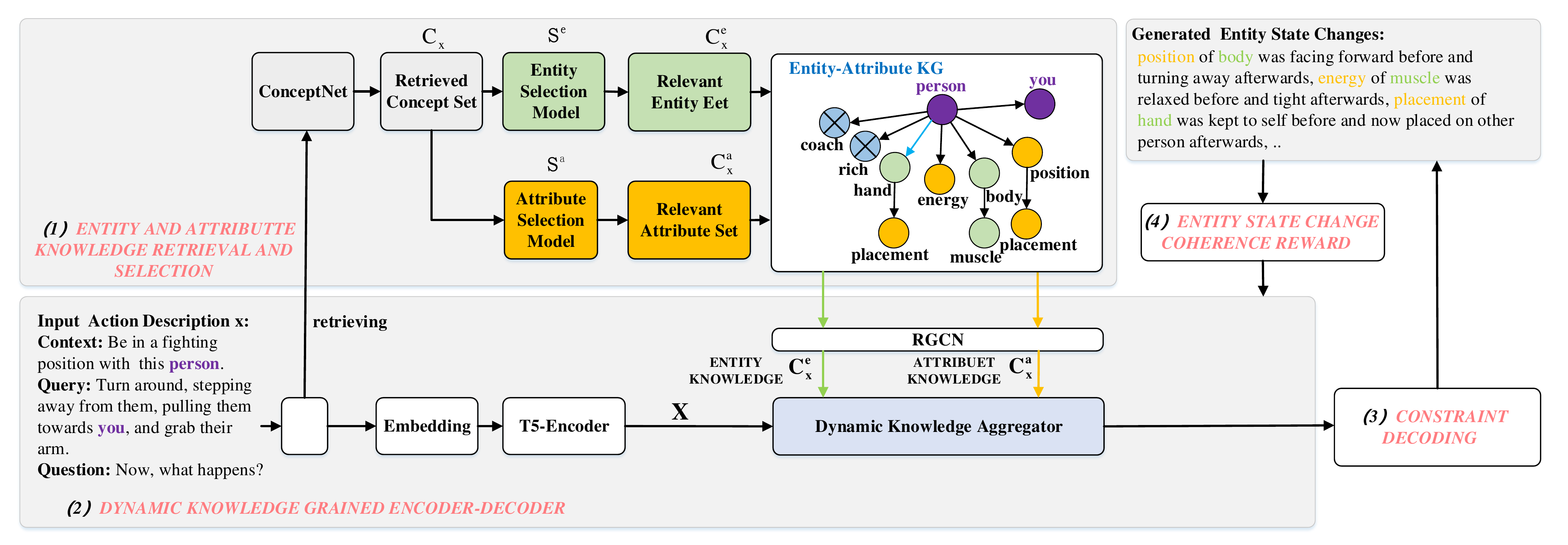}
 \vspace{-2mm}
	\caption{ Overview of the KIEST. In the Entity-attribute KG, the  green circle, orange circle, and blue circle  refer to the entity, attribute and noise entity separately. The black line and blue line  refer to the  relation \textit{relatedTo} and  \textit{capableof} separately.
 }
	\label{fig:whole framework-LIEST}
\end{figure*}

\subsection{Entity State Tracking} 
Tracking the state changes of entities is important to understand the natural language text describing actions and procedures. Most previous studies~\cite{bosselut2017simulating,mishra2018tracking,rashkin2018modeling,mysore2019materials,faghihi2021time,ma2022coalescing,singh2022frustratingly}
only focus on a particular domain with a set of predefined entities and attributes. For example, Mishra et al.~\cite{mishra2018tracking}
tackle this problem as a question-answering task and only focus on \textit{location} and \textit{existence} attributes to track the entity states in process paragraphs. Faghihi and Kordjamshidi~\cite{faghihi2021time} 
propose a Time-Stamped Language Model to understand the \textit{location} changes of entities. 
PiGLET~\cite{zellers2021piglet} predicts the post-state of an entity by giving its pre-state, a specific attribute and context, while EVENTS REALM~\cite{spiliopoulou2022events} determines whether an entity has a state change with respect to the set of given attributes. Recently, \cite{tandon2020dataset} further extended entity state tracking to open domain actions and explore pre-trained language models, such as GPT-2~\cite{radford2019language}, to autoregressively generate all the entity state changes. Compared with all these studies, our work focuses on open-domain entity state tracking and aims to encourage the model to generate entity state changes with high coverage and are more coherent to the context and human commonsense.

\subsection{Knowledge Informed Language Understanding and Generation}
Many studies \cite{velivckovic2017graph,sun2019ernie,liu2020k,sun2020colake,zhang2020minimize,wang2021kepler,al2021employing,andrus2022enhanced,ju2022grape,li2022hierarchical} have been proposed to incorporate external knowledge to better understand the text or generate the expected output.
For example, on the language understanding task, \cite{liu2020k} injects expanded knowledge into the language model by adding the entity and relation from  the knowledge graph as additional words.  Different from the masking strategy of BERT~\cite{devlin2018bert}, \cite{sun2019ernie}  proposes an entity-level masking  strategy to incorporate the informative entities into the language model. 
\cite{andrus2022enhanced} verbalize extracted facts aligning with input questions as natural language and incorporate them as prompts to the language model to improve story comprehension. 
For open-domain question answering,
\cite{ju2022grape} incorporate the informative entities extracted from the input question and passage with  the output of language model T5 to jointly optimize the knowledge representations based on their proposed relation-aware GNN.  
Inspired by these studies, we retrieve the relevant entities and attributes to the action description from external knowledge graphs and further dynamically incorporate them to better predict the entity state changes.

\section{Methodology}


Given a procedural paragraph with a sequence of action descriptions, we aim to predict all the state changes of entities related to each action. We follow~\cite{tandon2020dataset} and formulate the task of open domain entity state tracking as follows. The \textbf{Input} $x=(x_c,x_q)$
consists of a \textit{context} $x_c$
that includes one or a few history action descriptions and a \textit{query} $x_q$ that is the concatenation of a description for the current action and a short question ``\textit{what happens?}'', and the \textbf{Output} is a set of entity state changes $\mathbf{s}=\{s_j\}$ while each $s_j$ follows a template: [\textit{attribute}] \textbf{of}  [\textit{entity}] \textbf{was}  [\textit{before\_state}] \textbf{before and}  [\textit{after\_state}] \textbf{afterwards}. 


Intuitively, when asked to write state changes given an action description, humans will always call to mind the scenarios based on the action description, and then conceive the relevant entities and their state changes about certain attributes. We thus follow this human cognitive process and propose \textsc{Kiest}, a knowledge-informed framework to track the entity states given open domain actions. Specifically, as illustrated in Figure~\ref{fig:whole framework-LIEST}, \textsc{Kiest} consists of two main steps: (1) it first retrieves and selects all the candidate entities and attributes that are related to the action description from an external knowledge graph. Here, we use ConceptNet~\cite{liu2004conceptnet} given its high coverage of open-domain concepts; (2) it then dynamically incorporates the relevant entity and attribute knowledge into a \textbf{Dynamic Knowledge Grained Encoder-Decoder} to generate all the entity state changes while a constrained decoding strategy and a \textbf{Entity State Change Coherence Reward} are employed to encourage the decoder to generate state changes that are more coherent to the action descriptions and human commonsense. Next, we provide details for each of the components.

\subsection{Entity and Attribute Knowledge Retrieval and Selection}
\label{Entity-attribue rs}

We observe that state changes usually happen to the entities that are either directly included in the action description or are conceptually relevant to the key entities and actions (called as \textit{anchors}) included in the action description. For example, in Figure~\ref{fig:whole framework-LIEST}, the state change happens to the entity \textit{body}, which is closely related to the anchor \textit{person} in the action description. Similarly, the entity related attributes are also conceptually related to the anchors contained in the action description.


Motivated by this, we propose to acquire a rich set of entities and attributes that are relevant to state changes from ConceptNet~\cite{speer2017conceptnet}, a general KG covering about 4,716,604 open-domain concepts and their relations. Specifically, given the input $x=(x_c, x_q)$, we first find all the spans in $x$ that are included as concepts in ConceptNet and take each span as an anchor to retrieve the connected concepts within $H$ hops in ConceptNet, denoted as $C_{x}$. Taking the action description in Figure~\ref{fig:whole framework-LIEST} as an example, given the context and query as input, we first extract the anchors from them, such as \textit{person} and \textit{you}, and then take each anchor as a query to ConceptNet and obtain a set of relevant concepts, such as \textit{coach}, \textit{rich}, \textit{hand}, \textit{placement}, \textit{energy}, \textit{body}, \textit{position}, \textit{muscle} and so on.



For each input $x$, there are hundreds or thousands of neighboring concepts retrieved in $C_x$ while most of them are not related to entity state changes. For instance, in Figure~\ref{fig:whole framework-LIEST}, the retrieved  concepts, such as \textit{rich} and \textit{coach}, are not relevant to any state changes. Based on this observation, we further design two selection models $S^e$ and $S^a$ to select the most relevant entity and attribute knowledge to the action description from $C_x$. Both selection models share the same architecture and training objective. 
Taking $S^e$ as an example, it takes the following information as input: (1) action description $x$; (2) positive entity set $C^p=\{e^p\}$,
which is constructed based on the entities included in the human-annotated state changes; (3) negative entity set $C^n=\{e^n\}$, which is created by randomly sampling the entities from state changes annotated for other actions while ensuring each $e^n_v$ is not included in $C^p$.
To differentiate the positive and negative entities, $S^e$ utilizes a pre-trained BERT model $f(.)$~\cite{devlin2018bert,li2022sskgqa} to extract a semantic representation $f(x)$, $f(e^p)$, $f(e^n)$ for  $x$, $e^p$, $e^n$, respectively, and select the positive entities by measuring their distance to the action description. $S^e$ is optimized with the following triple loss:
\begin{align}
\max(\|f(x)\!\!-\!\!f(e^p)\|\!-\!\|f(x)\!\!-\!\!f(e^n)\|\!+\!\alpha,\!0) \nonumber
\end{align}
where $\|.\|$ denotes the Euclidean Distance and $\alpha$ is a margin parameter, which we set to 1 as default. During training, the triplet loss reduces the distance
between $f(x)$ and $f(e^p)$ while enlarging the
distance between $f(x)$ and $f(e^n)$. At inference time, we calculate the similarity scores between $x$ and each candidate concept from $C_x$ and select the candidates as relevant entity knowledge $C_x^{e}$ if their scores are higher than a threshold $\epsilon$. 
$\epsilon$ is regarded as a hyper-parameter and is discussed in section \ref{hyper-parameters-discuss}.

We utilize the same architecture for the attribute selection model $S^a$ and the same way to obtain the positive and negative set of attributes for each input $x$ to optimize $S^a$ with the triplet loss. 
Note that, during inference, for each input $x$, we apply $S^a$ to select the relevant attribute knowledge $C_x^{a}$ from 
the same set of candidate concepts $C_x$. 
After selecting the relevant entities and attributes for input $x$, we construct a heterogeneous knowledge graph, named entity-attribute knowledge graph (KG): $\mathcal{G}=(\mathcal{E}^x, \overline{C_x^{e}}, \overline{C_x^{a}}, \mathcal{R})$, where $\mathcal{E}^x$ denotes the set of anchor concepts extracted from the input $x$ and $\mathcal{R}$ denotes the links among the concepts of $\mathcal{E}^x$, $C_x^{e}$ and $C_x^{a}$. We remove a concept from $C_x^{e}$ and $C_x^{a}$ if it's not connected with any other concepts,
thus $\overline{C_x^{e}}$ and $\overline{C_x^{a}}$ are subsets of $C_x^{e}$ and $C_x^{a}$, respectively.

\begin{figure}[t]
	\centering
	\includegraphics[width=1\columnwidth]{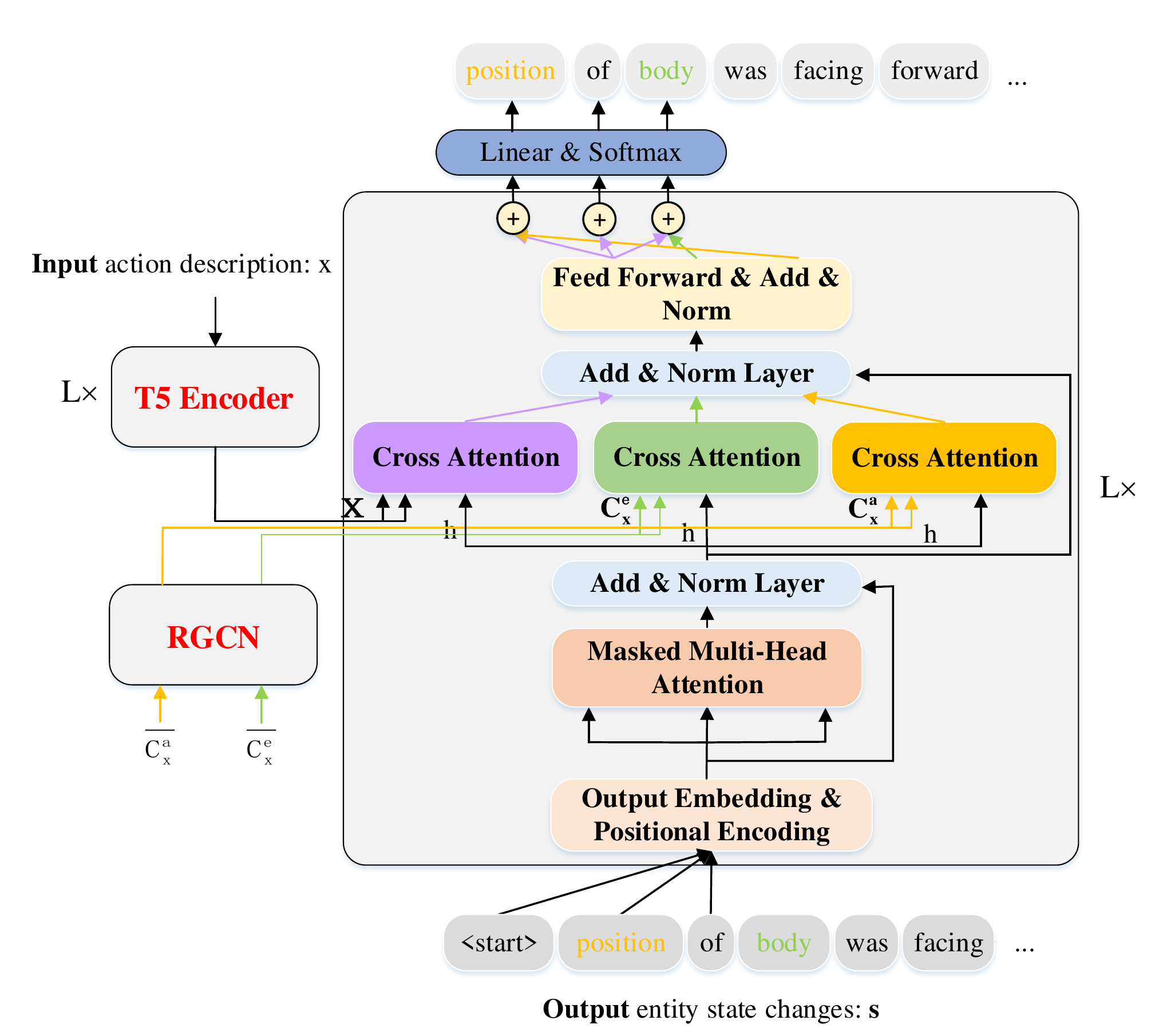}
	\caption{Overview of the DKGED decoder which dynamically incorporates additional relevant entities or attributes to generate entity state changes.
 }
	\label{fig:whole DKA}
\end{figure}

\subsection{Dynamic Knowledge Grained Encoder-Decoder Framework}

After selecting entities and attributes from ConceptNet that are relevant to the entity state changes, we aim to further incorporate them to improve the generation of the entity state changes. To do this, we propose a Dynamic Knowledge Grained Encoder-Decoder (DKGED) framework to dynamically select the relevant entity or attribute knowledge when we predict the tokens of each slot in the state change template, i.e., ``[\textit{attribute}] of [\textit{entity}] was [\textit{before\_state}] and [\textit{after\_state}] afterwards''. 
Specifically, DKGED takes the action description $x$, relevant entity set $\overline{C_x^{e}}$ and attribute set $\overline{C_x^{a}}$ as input and generates a set of entity state changes $\mathbf{s}=\{s_j\}$. We illustrate the DKGED at the bottom of Figure~\ref{fig:whole framework-LIEST} and describe its major components as follows.

DKGED first computes the hidden representation $\mathbf{X}=[\mathbf{w}_1,...,\mathbf{w}_n]$ of each token $w_n$ in the raw action description $x$: 
$$ 
\mathbf{X}=Encoder(w_1,...w_n)
$$
where $Encoder(.)$ is a $L$-layer Transformer encoder of T5. To capture the interaction among the relevant entities and attributes in the entity-attribute KG, we utilize the Relational Graph Convolutional Network (RGCN)~\cite{schlichtkrull2018modeling} to learn the embedding matrix $\mathbf{C_x^{e}}$ and $\mathbf{C_x^{a}}$ 
 for the entity set $\overline{C_x^{e}}$ and attribute set $\overline{C_x^{a}}$, respectively:
\begin{equation*}
\mathbf{C_x^{e}},\mathbf{C_x^{a}}=RGCN(\overline{C_x^{e}}, \overline{C_x^{a}})    
\end{equation*}

After encoding the input action description and the relevant entity and attribute knowledge, DKGED further autoregressively generates a set of entity state changes, denoted as a concatenation of a sequence of words $[\mathbf{s}_1,...,\mathbf{s}_i]$, while dynamically incorporating the 
relevant entities or attributes. Specifically, as shown in Figure~\ref{fig:whole DKA}, the DKGED adopts a $L$-layers of Transformer-based decoder, which converts the previous tokens into vector representations together with positional embedding and encodes them with a masked multi-head attention (MHA) in each layer:
$$\tilde{h}_i^l=\text{LN}(\text{MHA}(h_{1:i-1}^{l})+h_{i}^{l-1})$$
where $h_{1:i-1}^{l}, \tilde{h}_i^l$ denote the vector representation of previous tokens and current token at layer $l$, respectively. $h_{i}^{l-1}$ is the representation of the current token at layer $l-1$.   $\text{LN}$ denotes the Layer Normalization layer.

At each generation step $i$, DKGED dynamically incorporates the relevant entity knowledge, attribute knowledge, and the input action description to predict the output token, based on the position of $i$ in the state change template: ``[\textit{attribute}] of [\textit{entity}] was [\textit{before\_state}] and [\textit{after\_state}] afterwards''. Intuitively, when the current decoding step aims to predict an attribute, the external attribute knowledge is more meaningful than entity knowledge. Similarly, when the current decoding step is to predict an entity, the external entity knowledge is more helpful. Other positions, such as the template tokens (e.g., \textit{of}, \textit{was}, \textit{afterwards}) and states can be decoded based on the input action descriptions. To achieve this goal, 
we design three Cross Attention mechanisms while each Cross Attention is the same as Multi-Head Attention defined in Transformer~\cite{vaswani2017attention}:
\begin{align} 
&\mathbf{A}_i^l=\text{LN}(\text{CA}(\tilde{h_i^l},\mathbf{C_x^{a}})+\tilde{h_i^l}) & & i\in\{[e^b,e^e]\}
\nonumber\\
&\mathbf{E}_i^l=\text{LN}(\text{CA}(\tilde{h_i^l},\mathbf{C_x^{e}})+\tilde{h_i^l})  & & i\in\{[a^b,a^e]\}
\nonumber\\
&\mathbf{H}_i^l=\text{LN}(\text{CA}(\tilde{h_i^l},\mathbf{X})+\tilde{h_i^l}) & & \textit{other}
\nonumber
\end{align}
where $\mathbf{A}_i^l$, $\mathbf{E}_i^l$, and $\mathbf{H}_i^l$ is a contextual representation by attending over all the relevant attribute knowledge, entity knowledge, and all the input representations, respectively. $e^b$ and $e^e$ are the start and end word positions of the entity in the template. $a^b$ and $a^e$ are the start and end word positions of the attribute in the template. During the decoding process, these positions are known based on the previously generated template tokens, such as \textit{of}, \textit{was}.

Based on these contextual representations and the decoding position $i$ in the state change template, we further apply a Feed Forward layer to obtain an overall feature representation $h_i^l$. The feature representation from the last layer $h_i^L$ is then used to predict a probability distribution over the whole vocabulary $V$ based on Softmax
$$
P(\mathbf{s}_i|((\mathbf{s}_{1:i-1}, x, \overline{C_x^{e}}, \overline{C_x^{a}});\theta)=softmax(\mathbf{W}\cdot h^L_i) \nonumber
\label{main function}
$$
where $\mathbf{W}$ is a learnable parameter matrix.

\subsection{Constrained Decoding}
\label{cosntrained-decoding}
At each decoding step, existing studies~\cite{lewis2019bart,raffel2020exploring,hu2022planet} usually predict the output token from the whole vocabulary $V$, which results in outputs that are not related to the input. To improve the decoding process, we further design a \textit{constrained decoding} strategy to select a subset of candidate tokens that are more related to the input action description from the vocabulary $V$. Specifically, given an input $x$ and a candidate token $c$ from $V$, 
we calculate a \textit{cosine similarity score} between $c$ and each token in $x$ based on their contextual representations from a pre-trained T5 encoder, and use the highest score as the relevance score $\xi_x$ between $c$ and $x$:
\begin{align} 
&r^{}_n=cosine\_similarity(\mathbf{c},\mathbf{w}_n), w_n \in x \nonumber  \\
&\xi_x=\max_{}(r^{}_0, r^{}_1 ,..., r^{}_n) \nonumber
\end{align}
where $w_n$ is the $n-$th word in $x$,  $\mathbf{c}$ and $\mathbf{w}_n$ is the contextual representation of $c$ and $w_n$ from a pre-trained T5 encoder. For each input $x$, after computing the relevance scores for all the candidate tokens in $V$, we only select the tokens with relevance scores higher than $\gamma$, which is regarded as a hyper-parameter and discussed in section \ref{hyper-parameters-discuss}.

\subsection{Entity State Change Coherence Reward}
\label{subsec_esc}

An important issue with the generated state changes from baseline models such as GPT-2~\cite{radford2019language} is that the \textit{entity}, \textit{attribute}, \textit{before\_state} and \textit{after\_state} are not coherent enough or aligned with human commonsense. For example, in the generated output ``\textit{composition} of \textit{flowers} were \textit{in dead flowers} before and \textit{fresh flowers} afterwards'', the generated states, e.g., \textit{dead/fresh flowers}, are not relevant to the attribute \textit{composition}. To address this problem, we further design a classifier-based automatic metric to evaluate the coherence of each entity state change and use it as a reward to improve the generation process. To train the classifier, we use all the human-annotated entity state changes in the training set of OpenPI~\cite{tandon2020dataset} as the positive instances $S^p$, and create the same number of negative instances $S^n$ by randomly replacing the \textit{entity}, \textit{attribute}, \textit{before\_state}, \textit{after\_state} of each $s\in S^p$ with a concept sampled from other positive instances but the same slot. Then, each entity state change $s$ is fed as input to a T5 model which outputs a score ``1'' if the entity state change is coherent, otherwise, ``0''. We fine-tune the T5-based classifier with the same number of positive and negative training instances. The classifier is optimized with the cross-entropy objective:
\begin{align} 
p(l|s)=softmax(\text{T5\_Encoder}(s,\theta)) \nonumber
\end{align}
where $l$ is $0$ or $1$ and $p(l|s)$ is the probability indicating the coherence or incoherence of the entity state change $s$. $\theta$ denotes the set of parameters of the classifier. 

After training the classifier, we use it to estimate a coherence score for each generated entity state change and compute a reward based on the coherence score to further optimize the knowledge-grained encoder-decoder framework. 
The reward is computed by:


\begin{align} 
R_{cls}= log(p_s)(p(l_1|y)-p(l_0|y)) \nonumber
\end{align}
where $y$ is the generated target sequence sampled from the model’s distribution at each time step in decoding,
$p_s$ is the output probability of the sampled tokens. $l_0$ and $l_1$ denote the label 0 and 1, respectively. 
We then further optimize the DKGED framework with the reward using policy learning. The policy gradient is calculated by:
\begin{align} 
\nabla_{\phi}J(\phi)=E[R_{cls} \cdot\nabla_{\phi}log(P(y|x;\phi)) ] \nonumber
\end{align}
where $\phi$ represents the model parameters. $x$ is the action description.

The overall learning objective for our proposed \textsc{Kiest} is the combination of the RL and cross-entropy (CE) loss:
\begin{align}
&L_{CE}=-\sum_{i=1}^{k}log(\mathbf{s}_i|\mathbf{s}_{1:i-1},x,\overline{C_x^{e}},\overline{C_x^{a}})  \nonumber\\ 
&L_{RL}=-\frac{1}{t}\sum_{j=1}^{t}R_{cls}^{j}\nonumber\\ 
&L_{overall}=(1-\lambda)\cdot L_{CE} +\lambda \cdot L_{RL} \nonumber
\end{align} \label{eq:reward}
where $\lambda\in[0,1]$ is a tunable parameter.

\section{Experimental Setup}


\begin{table*}[ht]
	\centering
	\renewcommand\arraystretch{1.3}
\resizebox{0.7\textwidth}{!}{%
	\begin{tabular} {l|ccc|ccc|ccc}
		\toprule 
		\multicolumn{1}{c}  {}&\multicolumn{3}{c}  {Exact Match}& \multicolumn{3}{c}  {BLEU-2}& \multicolumn{3}{c}  {ROUGE-L}  \\
		 Approach & Precision &  Recall & F1 &  Precision &  Recall & F1 &  Precision &  Recall & F1  \\ 
	      \midrule
		GPT2-base~\cite{tandon2020dataset}&8.82 & 7.00&4.13 &22.85 &19.08 & 16.38& 39.54&35.18 &32.66\\
  GPT2-large~\cite{radford2019language}&10.82 & 7.69 &5.24  &25.35   & 20.59 &  18.16 & 41.88 &35.92  &34.08\\
 T5-base~\cite{raffel2020exploring}&20.10& 5.44&3.37&34.38&15.89 & 14.72& 49.86&29.95 &29.45\\
          T5-large~\cite{raffel2020exploring}& 15.22  & 7.46 & 5.64 & 31.30 &20.33 & 19.29& 47.83&35.97 &35.52\\
          \midrule
         \textsc{Kiest} w/o attribute &  19.83   &7.44   &6.66    &37.25   &21.79  & 21.68 &  52.73& 36.15 &  36.78\\
        \textsc{Kiest} w/o entity &19.41  &  7.30   &6.56   &36.95  & 21.69  &21.77   &52.74  &36.70  &37.36  \\
        \textsc{Kiest} w/o selection & 13.61 & 9.41   &6.26   &30.49      &\textbf{ 27.16}    &22.87   &47.19  &\textbf{ 42.88}  &39.29 \\
            \midrule
         \textsc{Kiest} w/o ESC+CD &  19.77  & 7.67 & 6.92  &37.66  &22.05 &22.17 &53.12 &36.63 &37.37 \\
        \textsc{Kiest} w/o ESC & \textbf{21.52}& 7.72   &  7.17&\textbf{ 39.73  }  &  22.53 & 22.92 &\textbf{54.77} & 37.10& 38.02\\
        \textsc{Kiest} (Our Approach) & 20.91& \textbf{9.54}  & \textbf{ 7.84 } & 38.14    & 26.29 &\textbf{ 24.24}  &53.43  & 40.77& \textbf{39.40}\\
           \bottomrule

	\end{tabular}
 }
\vspace{+2mm}
\caption{Results of various approaches for open domain entity state tracking on OpenPI.}
\label{con:Model main performance}
\vspace{-5mm}
\end{table*}
\subsection{Dataset}

We evaluate our approach on OpenPI~\cite{tandon2020dataset}, which, to the best of our knowledge, is the only public benchmark dataset for open domain entity state tracking. It comprises 23,880, 1,814, and 4,249 pairs of action description and entity state change in the training, development and test sets, respectively. We also design the following steps to correct the annotation errors of OpenPI:
\begin{itemize}
  \item  We first correct all the spelling errors contained in OpenPI, such as \textit{liqour} (``liquor''), \textit{skiier} (``skier''), \textit{necklacce} (``necklace''), \textit{apperance} (``appearance''), \textit{compostion} (``composition'') and so on. 
  
  \item some annotated entity state changes are not following the template ``[\textit{attribute}] of [\textit{entity}] was [\textit{before\_state}] and [\textit{after\_state}] afterwards'', such as ``\textit{flexibility of was hard before and soft afterwards}'', ``\textit{location of vegetable}''. We thus remove all these entity state changes from OpenPI. 
\end{itemize}
\subsection{Baselines and Evaluation Metrics}
We compare the performance of \textsc{Kiest} with several strong baselines based on the state-of-the-art pre-trained generative models, including \textbf{T5 base/large}~\cite{raffel2020exploring} and \textbf{GPT-2 base/large}~\cite{radford2019language}. Note that the only previous work~\cite{tandon2020dataset} on open domain entity state tracking utilized GPT-2-base \cite{radford2019language} to generate the state changes. We also design two variants of \textsc{Kiest} by removing the classifier-based coherence reward (denoted as \textsc{Kiest} w/o ESC), and removing both the classifier-based coherence reward and the constrained decoding strategy (denoted as \textsc{Kiest} w/o ESC+CD). Same as \cite{tandon2020dataset}, we evaluate all the models based on generative evaluation metrics, including Exact Match~\cite{Rajpurkar2016SQuAD10} 
, BLEU-2~\cite{papineni2002bleu}, and ROUGE-L~\cite{cohan2016revisiting}.

\subsection{Implementation Details}
In DKGED, the embedding of each node in   $\overline{C_x^{a}}$ or $\overline{C_x^{e}}$   is initialized  by the pre-trained model BERT \footnote{https://huggingface.co/bert-base-uncased} with 1024-dim. 
During training, we use AdamW~\cite{loshchilov2017decoupled} as the optimizer and Cross Entropy as the loss function with a learning rate of $0.00005$. We use label smoothing~\cite{muller2019does} to prevent the model from being over-confident. The batch size is 6.
For the overall loss function in section.\ref{subsec_esc}, we tune the $\lambda$ value in 0, 0.1, 0.3, 0.5, 0.7, 0.9. We find that \textsc{Kiest} works well when the weight of $L_{RL}$ is 0.1.
For the constrained decoding strategy, we tune the threshold $\gamma$ in $\{0, 0.2, 0.4, 0.6, 0.8\}$, and when $\gamma=0.4$, our approach achieves the best performance.   

In our entity selection model, we use AdamW Optimizer~\cite{loshchilov2017decoupled} with the learning rate of $0.00002$. We also use gradient clipping~\cite{zhang2019gradient} to constrain the max $L_2$-norm of the gradients to be 1. We tune the threshold $\epsilon$  in $\{0.3, 0.5\}$\footnote{we found there are  60\%, 52\%, 40\% entities in $C_x^e$ belongs to the training entity set when $\epsilon=0.3, 
 0.5, 0.7$ respectively. To obtain abundant knowledge, we discuss the model performance when  $\epsilon=0.3, 0.5$.
}, and hops $H$ in $\{1, 2\}$.  We find that our model works well when $\epsilon=0.5, H=2$. 
By evaluation, we use the same configuration as above in our attribute selection model. By using selection models, the maximum number of filtered entities and attributes in all action descriptions are  1000 and 140, so when training our model, the maximum number of filtered entities and attributes are set to 1000 and 140 in each action description.

To train the classifier for estimating the coherence of each entity state change, we set the dropout rate to 0.1 in the last linear layer, batch size to 32, and use AdamW optimizer~\cite{loshchilov2017decoupled} with the learning rate of $0.00002$. We also apply gradient clipping~\cite{zhang2019gradient} to constrain the maximum value of $L_2$-norm of the gradients to be 1.  The accuracy of predicting \textit{coherence} or \textit{incoherence} for the development and test sets of OpenPI is both 98\%, demonstrating that the performance of the classifier has been high enough to be used as a reward function.  


\section{Results and Discussion}
Table~\ref{con:Model main performance} presents the experimental results of various approaches based on the metrics, including Exact Match, BLEU-2 and ROUGE-L. We have the following observations: (1) our \textsc{Kiest} significantly outperforms all the strong baselines and its variants across all evaluation metrics; (2) T5-large model shows obvious superiority compared with GPT-2-base/large model. 
(3) by adding the classifier-based coherence reward, our \textsc{Kiest} approach significantly improves over the baseline \textsc{Kiest} w/o ESC, especially on recall of all evaluation metrics, demonstrating the effectiveness of the classifier-based coherence reward especially in encouraging the model to generate more valid entity state changes. The overall precision is dropped after adding the classifier-based coherence reward. We guess the reason is that \textsc{Kiest} tends to generate more entity state changes than \textsc{Kiest} w/o ESC. 
(4) the constrained decoding can help remove most of the noisy tokens from the target vocabulary, thus the precision of the approach is significantly improved, which can be seen by comparing 
\textsc{Kiest} w/o ESC+CD with \textsc{Kiest} w/o ESC).

\begin{table}[h]
	\centering
	\renewcommand\arraystretch{1.1}
	\begin{tabular} {l|cccc}
		  \midrule
    &  \multicolumn{2}{c}{Training} &\multicolumn{2}{c}{Inference}\\ 
     \midrule
		   & R-Time &S-Space&R-Time&S-Space\\ 	
               \midrule
		w/ selection &04:23:11  & 3,793 MB & 00:29:46 & 38,425 MB \\ 
		  \midrule
		w/o selection &71:30:21  & 4,845 MB & 00:38:09 & 40,112 MB \\
		  \midrule
	\end{tabular}
 \vspace{+2mm}
 	\caption{Reduction of storage and run time on OpenPI dataset.
  S-Space refers to Storage Space, R-Time refers to Running Time ( hours: minutes: seconds), w/o selection refers to without selection model,  w/ selection refers to with selection model.}
		\label{ref:time_comsuming}
  \vspace{-5mm}
\end{table}

Ablation studies based on different types of external knowledge and selection strategy are shown in the middle of Table \ref{con:Model main performance}. We have the following observations: (1) By comparing
the performance of only injecting the entity knowledge $\mathbf{C_x^e}$ (w/o attribute) or attribute knowledge $\mathbf{C_x^a}$ (w/o entity) \footnote{When injecting the entity/attribute knowledge, the generation of other positions only uses the knowledge $\mathbf{X}$.} to our model, our model still get a significant improvement over the best generation model T5-large, demonstrating that by incorporating the relevant entity or attribute knowledge, the generation of entity state changes can be significantly improved, which validates our motivation that the relevant entity and attribute knowledge from the knowledge graph can help improve the coverage of the generation of entity state changes. 
(2) Notably, without the entity and attribute selection (\textsc{Kiest} w/o selection), the precision of \textsc{Kiest} significantly drops on all evaluation metrics, which validates our assumption that without selection, the massive concepts retrieved from ConceptNet are likely to introduce noise and hurt the model's performance.

To evaluate the effectiveness of our proposed selection method in reducing storage and running time, we compared the storage space and run time between \textsc{Kiest} with and without selection on the OpenPI dataset. As presented in Table \ref{ref:time_comsuming}, our model without selection required 4,845 MB of storage whereas \textsc{Kiest} with selection only used 3,793 MB, resulting in a 23.6\% reduction in storage costs. This reduction is primarily due to the selection approach which reduced noise from the original entity and attribute sets. Furthermore, the selection method further significantly reduced the running time cost for both training and inference.

\section{Analysis}


\subsection{The Impact of Hyper-Parameters}
\label{hyper-parameters-discuss}
\begin{figure}[t]
        \centering
        \includegraphics[width=\columnwidth]{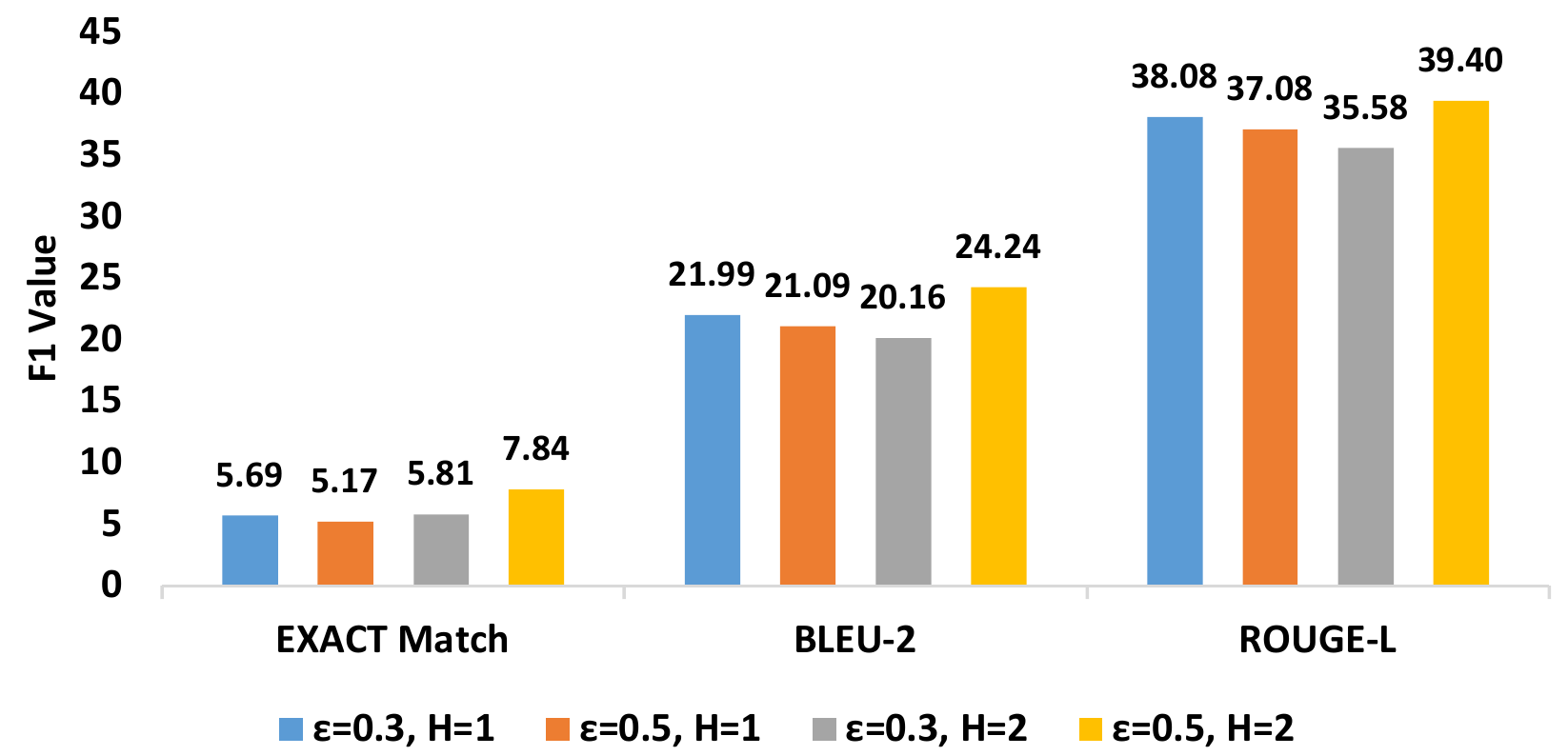}
	\caption{ F1  results with different  $\epsilon$ and $H$ settings. 
 }
	\label{paramters_discuss}
\end{figure}

In our proposed model, there are two parameters controlling the size of the relevant entities and attributes retrieved from the ConceptNet: (1) $\epsilon \in \{0.3, 0.5\}$ which is the threshold to select relevant entities and attributes based on their distance to the input action description, and (2) $H\in\{1, 2\}$ which is the number of hops to retrieve the concepts that are related to the anchors of the action description from ConceptNet. We analyze the impact of these two hyper-parameters in Figure~\ref{paramters_discuss}. As we can see, when $\epsilon=0.5$, \textsc{Kiest} selects entities and attributes that are more relevant to the input action description and consistently provides better performance than the setting of $\epsilon=0.3$. In addition, $H=1$ is also better than $H=2$, indicating that when we set $H=2$, too many concepts are retrieved from ConceptNet which leads to too much noise and hurts the performance of both the selection models and \textsc{Kiest}.


\begin{figure}[t]
        \centering
        \includegraphics[width=\columnwidth]{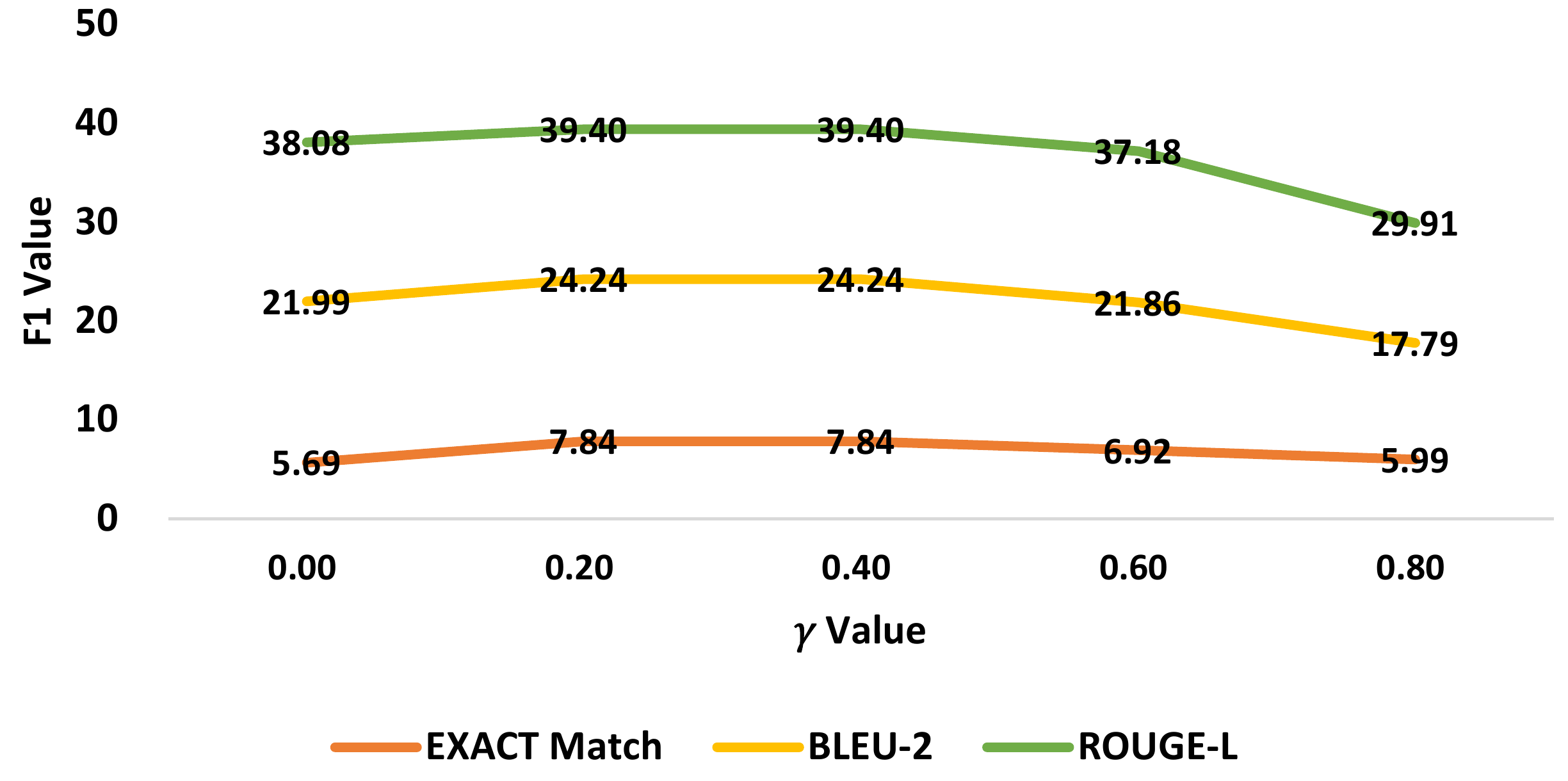}
	\caption{F1 results with different $\gamma$ settings.}
	\label{paramters_discuss_2}
\end{figure}

We also analyze the impact of $\gamma\in\{0, 0.2, 0.4, 0.6, 0.8\}$, which is the threshold to select tokens for the target output vocabulary. As Figure~\ref{paramters_discuss_2} shows, \textsc{Kiest} achieves the best performance when $\gamma$ is $0.2$ or $0.4$. When $\gamma<0.2$, too many candidate tokens are considered at each decoding step and thus leading to irrelevant output. When $\gamma>0.4$, too few candidate tokens are included in the target vocabulary and thus also leading to negative impacts on the generation of entity state changes. 

\subsection{Content Richness and Coherence of Outputs}

\begin{figure}[t]
        \centering
        \includegraphics[width=\columnwidth]{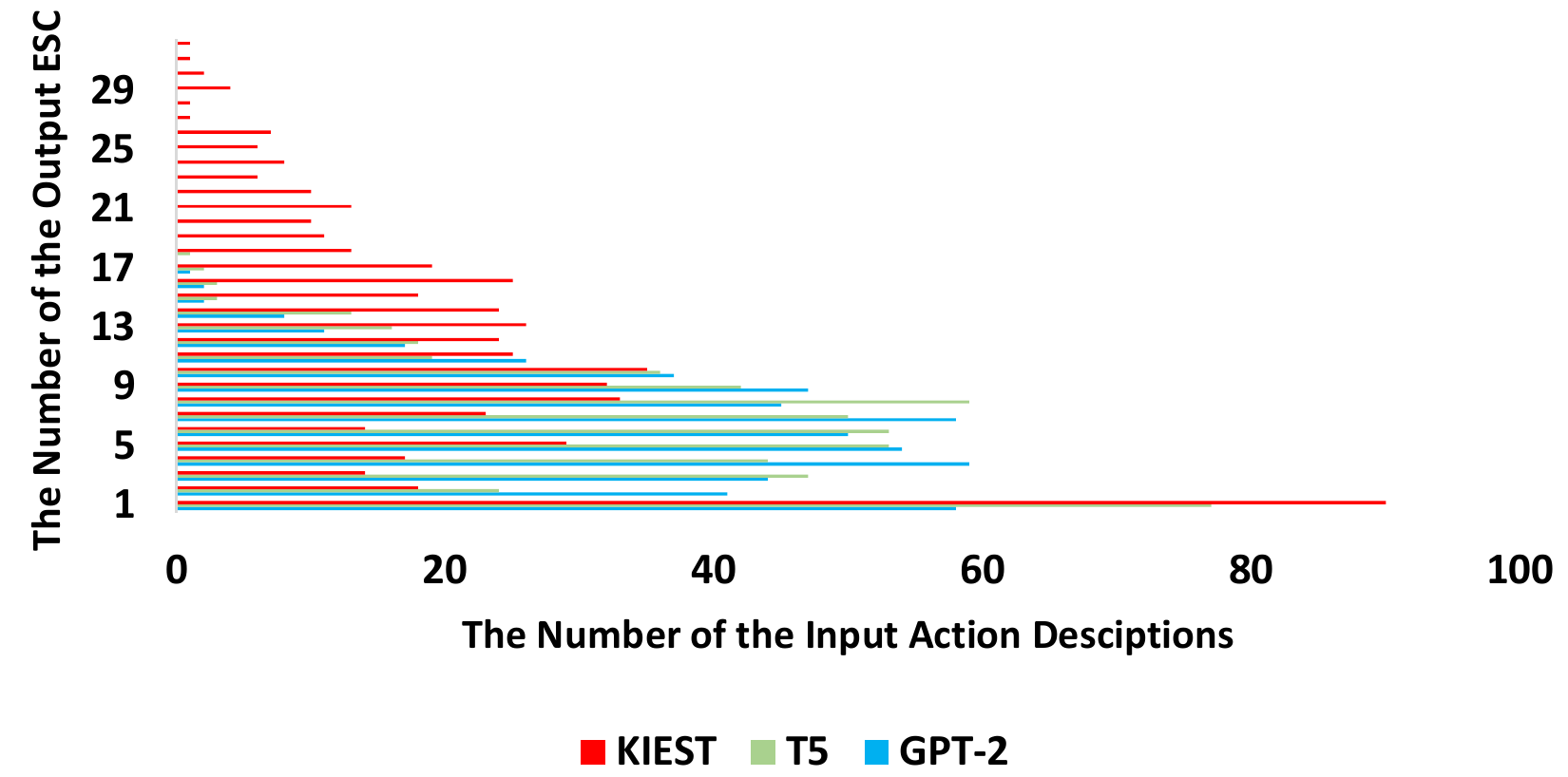}
	\caption{Number of entity state changes for each action description generated by different approaches.}
	\label{number-of-esc}
\end{figure}

We measure the content richness of \textsc{Kiest} based on the generated entity state changes and compare it with the two strong baseline models, GPT-2-Large and T5-Large. As shown in Figure~\ref{number-of-esc}, \textsc{Kiest} tends to generate longer outputs, e.g.,  more entity state changes with sequence length in the range of 16-31 than the baseline models. 
The results imply that our knowledge-informed framework encourages the model to generate more entity state changes with higher coverage.

\begin{figure}[t]
        \centering
        \includegraphics[width=\columnwidth]{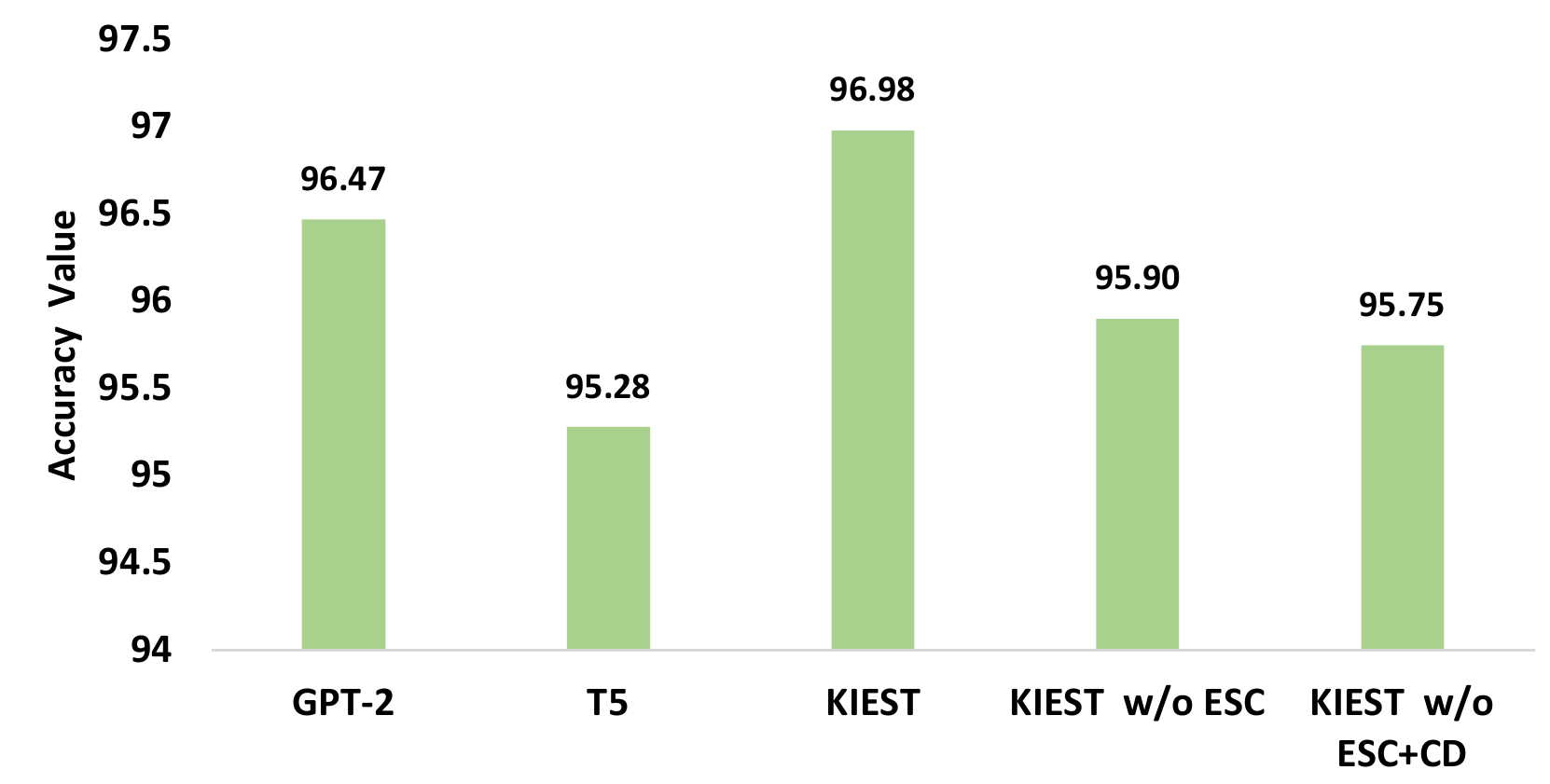}
	\caption{ Automatic coherence evaluation of different models.}
	\label{coverage-and-coherent_1}
\end{figure}

To evaluate the coherence of the entity state changes generated by \textsc{Kiest}, its variants and the baseline models, we use the trained classifier in Section~\ref{subsec_esc} as an automatic metric. The output of the classifier is normalized into $[0,1]$ with the sigmoid function. 
As shown in Figure~\ref{coverage-and-coherent_1}, \textsc{Kiest} achieves the highest coherence compared with the GPT-2 and T5, suggesting that 
our knowledge-informed model with the constraint decoding and coherent reward method is effective to improve the coherence of entity state change.
For further analysis,
\textsc{Kiest} w/o ESC+CD has a lower score compared to GPT-2, suggesting that simply incorporating the additional entity and attribute knowledge into the language model but without any control on the decoding process may have a negative impact on the coherence of the output. Without using the classifier-based coherence reward (i.e., \textsc{Kiest}  w/o ESC), the average coherence score of the outputs is lower than the coherence score of \textsc{Kiest}, demonstrating the effectiveness of the coherence reward.

\subsection{Is ChatGPT a Good Open Domain Entity State Tracker?}
Recently, ChatGPT from OpenAI has shown significant advances in various downstream NLP tasks. Here, we systematically compare the state-of-the-art GPT models, including 
GPT-3.5-turbo\footnote{https://platform.openai.com/docs/models/gpt-3-5}, GPT-4\footnote{https://platform.openai.com/docs/models/gpt-4}, with \textsc{Kiest} on open domain entity state tracking task, to answer the potential research question: "Is ChatGPT a good open domain entity state tracker?".

\begin{figure*}[t]
        \centering
        \includegraphics[width=1.8\columnwidth]{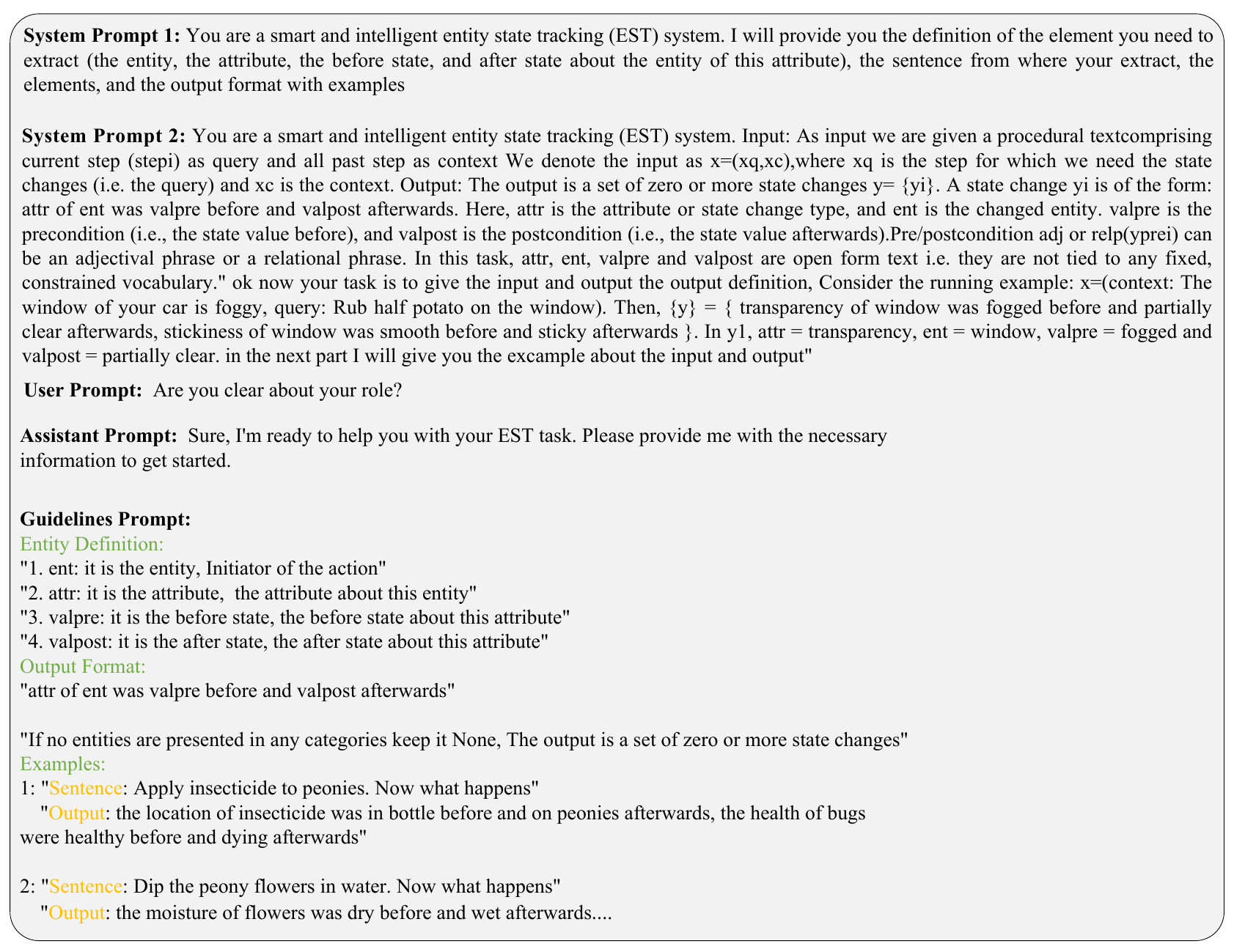}
	\caption{ Example of Prompt 1 and Prompt 2 defined for GPT-based models.}
	\label{prompt}
\end{figure*}
We first design prompts to instruct the GPT models to predict the entity state changes for each input action description. As shown in Figure~\ref{prompt}, we design two prompts: (1) Prompt 1 is based on a task description defined manually; (2) prompt 2 contains a task definition from~\cite{tandon2020dataset} and 8 demonstration examples that are randomly selected from the OpenPI training dataset. 
Table~\ref{ref:GPT-3.5_results} presents the zero-shot performance of GPT-3.5-turbo and GPT-4 on the whole testing set of OpenPI. Our analysis reveals that different prompts yield significantly different results for both GPT-3.5-turbo and GPT-4. 
Although GPT-4 outperforms GPT-3.5-turbo in terms of Exact Match and BLEU-2 scores, both models fail to achieve satisfactory results on this task compared to all the supervised models including \textsc{Kiest}. These findings suggest that, without any tuning, the large language models still have limited capability to perform complex reasoning tasks.
\begin{table}[h]
	\centering
	\renewcommand\arraystretch{1.1}
	\begin{tabular} {l|c|c|c}
	\midrule
		 Model& Exact Match &BLEU-2& ROUGE-L\\
    \midrule
        GPT-3.5-turbo (P1)&0.34 &  6.74 &  21.25   \\
        GPT-3.5-turbo (P2)&0.38 &  8.28   &   24.06    \\
        GPT-4 (P1)& 1.25&  11.02 & 27.28  \\
       GPT-4 (P2) & 0.56&  10.34 & 26.78  \\
 \textsc{Kiest} & 7.84&  24.24 & 39.40  \\
     \midrule
	\end{tabular}
  \vspace{+2mm}
  \caption{F1 performance of GPT-3.5-turbo and GPT-4 on OpenPI. P1 refers to Prompt1, and P2 refers to Prompt2.}
  \vspace{-5mm}
  \label{ref:GPT-3.5_results}
\end{table}



\begin{table}[ht]
	\centering
		
	\renewcommand\arraystretch{1.3}
	\scalebox{0.8}{
	\begin{tabular} {lll}
		\hline 
		\textbf{Model}&\multicolumn{2}{l}{\textbf{Wrong Template Prediction:}}  \\ 
		\hline

		\multirow{2}*{ GPT-3.5-turbo (P1)}&Action Input:&\makecell[l]{Identify the state where the car is\\ licensed. Now, what happens?}\\
		~&\textcolor{teal}{Generated ESC:} & \textcolor{teal}{\makecell[l]{ent is car, attr is licensed state,valpre \\is unknown and valpost is the state \\where the car was licensed}} \\
\hline	
        \multirow{2}*{ GPT-3.5-turbo (P2)}&Action Input:&\makecell[l]{Identify the state where the car  is \\ licensed. Subscribe to an online \\ information  search website. \\Now, what happens?	}\\
		~&\textcolor{teal}{Generated ESC:} & \textcolor{teal}{\makecell[l]{access to information was limited before \\ and expanded afterwards.} }\\

        \hline

	\multirow{2}*{GPT-4 (P1)}&Action Input:&\makecell[l] {Identify the state where the car is \\licensed. Subscribe to an online \\ information search website. \\Now, what happens?}\\
		~&\textcolor{teal}{Generated ESC: }& \textcolor{teal}{\makecell[l]{information search website subscription \\ status was inactive before and active \\afterwards} }\\
     \hline
      \hline
  \textbf{}&\multicolumn{2}{l}{\textbf{Unable to Extract Entities or Attributes:}}    \\ 
\hline
	\multirow{2}*{GPT-3.5-turbo (P1)}&Action Input:&\makecell[l] {Grind roots, seeds and other hard \\ herbs in a  coffee grinder or mortar \\and  pestle. Mix the ground or dried\\ herb with  a bit of oil or water until it \\ forms a moist paste. Now, what happens?}\\
		~&\textcolor{teal}{Generated ESC:} & \textcolor{teal}{\makecell[l]{attr of None was None before and None \\ afterwards} }\\
   \hline
\multirow{2}*{GPT-3.5-turbo (P2)}&Action Input:&\makecell[l] {It needs to be SMART: specific, \\ measurable, attainable, realistic and \\timed. Identify the necessary \\steps. Achieving your goals can often \\ take a  long time and can be very \\frustrating. Now, what happens?}\\
		~&\textcolor{teal}{Generated ESC:} & \textcolor{teal}{\makecell[l]{None, as no entities with attributes\\ are presented in the given sentence.} }\\
   \hline

  	\multirow{2}*{GPT-4 (P2)}&Action Input:&\makecell[l] {Gather the supplies you will need:\\ Now, what happens?}\\
		~&\textcolor{teal}{Generated ESC: }&\textcolor{teal}{ \makecell[l]{There will be no change.}} \\
   \hline
      \hline
    \textbf{}&\multicolumn{2}{l}{\textbf{Generating Irrelevant Entity State Change:}}    \\ 
    \hline
	\multirow{2}*{GPT-3.5-turbo (P1)}&Action Input:&\makecell[l] {Do something that smells great, like \\bake bread. Now, what happens?}\\
		~&\textcolor{teal}{Generated ESC:} & \textcolor{teal}{\makecell[l]{state of \textbf{hunger} of people around \\ was not hungry before and potentially \\ hungry afterwards} }\\
 \hline
  	\multirow{2}*{GPT-4 (P1)}&Action Input:&\makecell[l] {Do something that smells great, \\like bake bread. Now, what happens?}\\
		~&\textcolor{teal}{Generated ESC:} & \textcolor{teal}{\makecell[l]{smell of \textbf{flowers} were not present \\ before  and present afterwards}} \\
 \hline
\multirow{2}*{GPT-4 (P2)}&Action Input:&\makecell[l] {Have a spa experience. Read a good \\ book. Now, what happens?}\\
		~&\textcolor{teal}{Generated ESC: }& \textcolor{teal}{\makecell[l]{knowledge of \textbf{plot book} was unknown\\ before and in \textbf{parent's mind} afterwards} }\\
\hline

\hline
	\end{tabular}}
	\caption{The examples of prediction error about GPT-3.5-turbo (P1/P2) and GPT-4 (P1/P2).}

\label{con:GPT_error_prediction}
\end{table}

We further manually analyze the prediction errors of GPT-3.5-turbo and GPT-4 on the test set of OpenPI and summarize them into the following categories (examples for each category are shown in Tabel~\ref{con:GPT_error_prediction}):
\begin{itemize}
  \item \textbf{Wrong Template Prediction:} Despite being provided with output templates and examples, GPT-3.5-turbo and GPT-4 still generated a considerable number of output errors that are not based on the provided template. 
  Furthermore, our analysis indicates that GPT-3.5-turbo is more prone to generating erroneous templates than GPT-4.
  \item \textbf{Unable to Extract Entities or Attributes:} Our analysis indicates that GPT-3.5-turbo and GPT-4 struggle to comprehend the semantic information of input action descriptions and accurately identify the corresponding entities, attributes, and states. 
  \item \textbf{Generating Irrelevant Entity State Changes:} 
    During our analysis, we observed that GPT-3.5-turbo and GPT-4 occasionally generate irrelevant Entity State Changes. As an illustration in Tabel~\ref{con:GPT_error_prediction},
    these examples demonstrate that GPT-3.5-turbo and GPT-4 are capable of generating semantically plausible entity state changes, but in some cases, they may not be aligned with the information presented in the input action description. Such as for the generations of model GPT-3.5-turbo (P1) and GPT-4 (P1), the entities ``\textit{hunger}'' and ``\textit{flowers}'' in the generated output are not relevant to the context described in the input.
\end{itemize}

\section{Limitation}
In this work, we explored various state-of-the-art large language models for the open-domain entity state-tracking task. Though such models are capable of generating relevant and fluent outputs given the input action description, they still cannot generate all the accurate entity state changes with high coverage. In our proposed \textsc{Kiest} framework, we attempted to locate the relevant entities and attributes from ConceptNet. Such knowledge is still noisy and the coverage is not satisfying. Going forward, we plan to develop more appropriate prompts to stimulate and instruct the language models such as GPT-3.5 or GPT-4 to generate more coherent entity state changes. In addition, this work only considers textual descriptions of actions while their corresponding image or video illustrations are also beneficial for predicting the relevant entities and state changes. In the future, we will extend entity state tracking to multimodality by leveraging the state-of-the-art multimodal pre-trained language models~\cite{wang2022ofa} and instruction tuning~\cite{xu2022multiinstruct}.



\section{Conclusion}

This paper introduces \textsc{Kiest}, a knowledge-informed framework for open domain entity state tracking. It consists of two major steps: first, retrieving and selecting the relevant entity and attribute knowledge from an external knowledge base, i.e., ConceptNet; and then dynamically incorporating the entity and attribute knowledge into an encoder-decoder framework with an effective constrained decoding strategy and a classifier-based entity state change coherence reward. Experimental results and extensive analysis on the public benchmark dataset -- OpenPI demonstrate the effectiveness of our overall framework and each component with significant improvements over the strong baselines, including T5, GPT-2, GPT-3.5, and GPT-4.



\section*{Acknowledgements}
This research is based upon work supported by U.S. DARPA KMASS Program \# HR001121S0034. The views and conclusions contained herein are those of the authors and should not be interpreted as necessarily representing the official policies, either expressed or implied, of DARPA or the U.S. Government. The U.S. Government is authorized to reproduce and distribute reprints for governmental purposes notwithstanding any copyright annotation therein.


\bibliographystyle{ACM-Reference-Format}
\balance
\bibliography{9_sample-base}

\end{document}